\documentclass[twoside,twocolumn]{article}

\usepackage{blindtext} % Package to generate dummy text throughout this template 

\usepackage{authblk}

\usepackage[sc]{mathpazo} % Use the Palatino font
\usepackage[T1]{fontenc} % Use 8-bit encoding that has 256 glyphs
\usepackage{microtype} % Slightly tweak font spacing for aesthetics

\usepackage[english]{babel} % Language hyphenation and typographical rules

\usepackage[hmarginratio=1:1,top=32mm,columnsep=20pt]{geometry} % Document margins
\usepackage[hang, small,labelfont=bf,up,textfont=it,up]{caption} % Custom captions under/above floats in tables or figures
\usepackage{booktabs} % Horizontal rules in tables

\usepackage{lettrine} % The lettrine is the first enlarged letter at the beginning of the text

\usepackage{enumitem} % Customized lists
\setlist[itemize]{noitemsep} % Make itemize lists more compact

\usepackage{abstract} % Allows abstract customization
\usepackage{titlesec} % Allows customization of titles
\renewcommand\thesection{\Roman{section}} % Roman numerals for the sections
\renewcommand\thesubsection{\roman{subsection}} % roman numerals for subsections
\titleformat{\section}[block]{\large\scshape\centering}{\thesection.}{1em}{} % Change the look of the section titles
\titleformat{\subsection}[block]{\large}{\thesubsection.}{1em}{} % Change the look of the section titles

\usepackage{fancyhdr} % Headers and footers
\usepackage{hyperref} % For hyperlinks in the PDF

\usepackage{graphicx}
\usepackage{subcaption}

\title{\vspace{-1.7cm}\makebox[0pt]{FieldSAFE: Dataset for Obstacle Detection in Agriculture}} %Agricultural Dataset with Static and Moving Obstacles} % Article title
\date{} % Leave empty to omit a date %\today

\author[1]{Mikkel Kragh\thanks{\{mkha,pech\}@eng.au.dk\\ \indent Both authors contributed equally to this work.}}
\author[1]{Peter Christiansen$^{*}$}
\author[1]{Morten S. Laursen}
\author[2]{Morten Larsen}
\author[3]{Kim A. Steen}
\author[3]{Ole Green}
\author[1]{Henrik Karstoft}
\author[1]{Rasmus N. J{\o}rgensen}
\affil[1]{Department of Engineering, Aarhus University, Denmark}
\affil[2]{Conpleks Innovation ApS, Struer, Denmark}
\affil[3]{AgroIntelli, Aarhus, Denmark}

\providecommand{\keywords}[1]{\textbf{\textit{Keywords ---}} #1}

\begin{document}
% Print the title
\maketitle

% Change abstract settings to no margin
\renewenvironment{abstract}
 {\small
  \begin{center}
  \bfseries \abstractname\vspace{-.5em}\vspace{0pt}
  \end{center}
  \list{}{%
    \setlength{\leftmargin}{0mm}% <---------- CHANGE HERE
    \setlength{\rightmargin}{\leftmargin}%
  }%
  \item\relax}
 {\endlist}

\begin{abstract}
\noindent
In this paper, we present a novel multi-modal dataset for obstacle detection in agriculture.
The dataset comprises approximately 2 hours of raw sensor data from a tractor-mounted sensor system in a grass mowing scenario in Denmark, October 2016.
Sensing modalities include stereo camera, thermal camera, web camera, 360-degree camera, lidar, and radar, while precise localization is available from fused IMU and GNSS.
Both static and moving obstacles are present including humans, mannequin dolls, rocks, barrels, buildings, vehicles, and vegetation.
All obstacles have ground truth object labels and geographic coordinates.
%Ground truth maps with precise GPS coordinates are available for both static and dynamic obstacles.
\end{abstract}
%}

%----------------------------------------------------------------------------------------
%	ARTICLE CONTENTS
%----------------------------------------------------------------------------------------

\begin{figure}[t]
    \centering
    \includegraphics[width=0.5\textwidth]{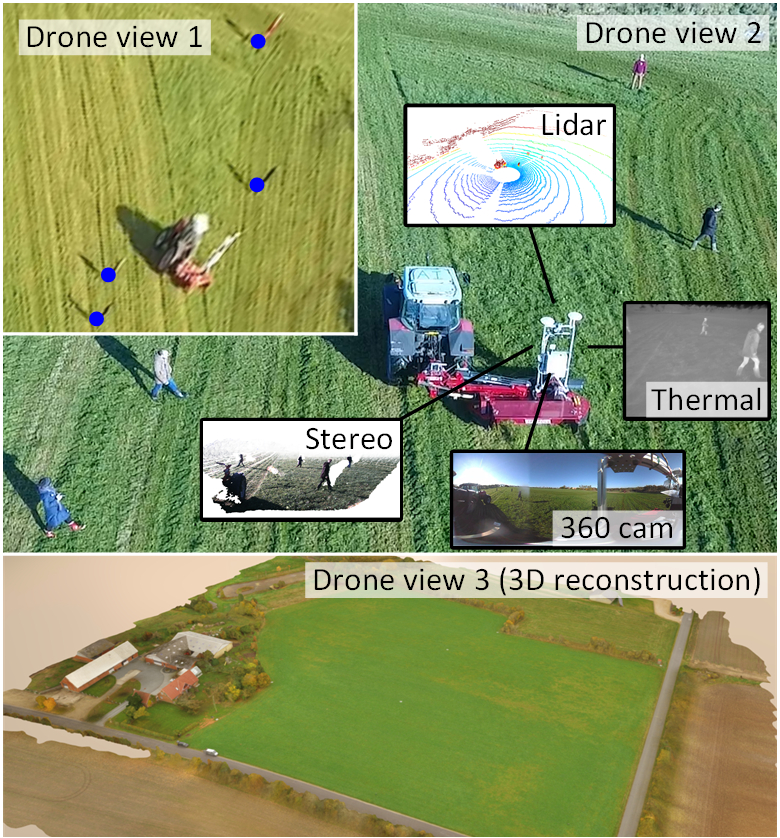}
    \caption{Recording platform surrounded by static and moving obstacles. Multiple drone views record the exact position of obstacles, while the recording platform records local sensor data.}\label{fig:teaser}
    %\vspace*{-0.5cm}
\end{figure}

\section{Introduction}

For the past few decades, precision agriculture has revolutionized agricultural production systems.
Part of the development has focused on robotic automation, to optimize workflow and minimize manual labor.
Today, technology is available to automatically steer farming vehicles such as tractors and harvesters along predefined paths using accurate global navigation sattelite systems (GNSS) \cite{Abidine2004}.
However, a human is still needed to monitor the surroundings and intervene when potential obstacles appear in front of the vehicle to ensure safety.

In order to completely eliminate the need for a human operator, autonomous farming vehicles need to operate both efficiently and safely without any human intervention.
A safety system must perform robust obstacle detection and avoidance in real-time with high reliability.
And multiple sensing modalities must complement each other in order to handle a wide range of changes in illumination and weather conditions.

A technological advancement like this requires extensive research and experiments to investigate combinations of sensors, detection algorithms, and fusion strategies.
Currently, a few R\&D projects exist within companies that seek to commercialize the concept \cite{CaseIH,AutonomousSolutions,Kubota}.
However, to our knowledge, no public platforms or datasets are available that address the important issues of obstacle detection in an agricultural environment.
%Before then, however, experiments and research are needed to investigate which detection algorithms, fusion strategies, and sensing combinations are needed.
%In the future, autonomous farming vehicles will operate both efficiently and safely without any human intervention.
%A safety system will perform robust obstacle detection and avoidance in real-time with high reliability.
%And multiple sensing modalities will complement each other in order to handle a wide range of changes in illumination and weather conditions.
%Before then, however, experiments and research are needed to investigate which detection algorithms, fusion strategies, and sensing combinations are needed.

%In recent years, a number of datasets addressing autonomous driving have been made publicly available.
Within urban autonomous driving, a number of datasets have recently been made publicly available.
Udacity's Self-Driving Car Engineer Nanodegree program has given rise to multiple challenge datasets including stereo camera, lidar, and localization data \cite{Didi1,Didi2,DidiChallenge}.
A few research institutions such as the University of Surrey \cite{DIPLECS}, Link\"{o}ping University \cite{amuse2013gt}, Oxford  \cite{RobotCarDatasetIJRR}, and Virginia Tech \cite{insight} have published similar datasets.
Most of the above cases, however, only address behavioural cloning, such that ground truth data are only available for control actions of the vehicles.
No information is thus available for potential obstacles and their location in front of the vehicles.

The KITTI dataset \cite{Geiger2013IJRR}, however, addresses these issues with object annotations in both 2D and 3D.
Today, it is the de facto standard for benchmarking both single- and multi-modality object detection and recognition systems for autonomous driving.
The dataset includes a high-resolution stereo camera, a 360-degree camera, a lidar, and fused GNSS/IMU sensor data.

Focusing specifically on image data, an even larger selection of datasets is available with annotations of typical object categories such as cars, pedestrians, and bicycles.
Annotations of cars are often represented by bounding boxes \cite{MatzenICCV13,TMEMotorwayDataset}.
However, pixel-level annotation or semantic segmentation has the advantage of being able to capture all objects, regardless of their shape and orientation.
Some of these are synthetically generated images using computer graphic engines that are automatically annotated \cite{RosCVPR16,GaidonVirtualCVPR2016}, whereas others are natural images that are manually labeled \cite{Cordts2016Cityscapes,MVD2017}.

In agriculture, currently no similar datasets are publicly available.
While some similarities between autonomous urban driving and autonomous farming are present, essential differences exist.
An agricultural environment is often unstructured or semi-structured, whereas urban driving involves planar surfaces, often accompanied by lane lines and traffic signs.
Further, distinction between traversable, non-traversable and processable terrain is often necessary in an agricultural context such as grass mowing, weed spraying, or harvesting.
Here, tall grass or high crops protruding from the ground may actually be traversable and processable, whereas ordinary object categories such as humans, animals, and vehicles are not.
In urban driving, however, a simplified traversable/non-traversable representation is common, as all protruding objects are typically regarded as obstacles.
Therefore, sensing modalities and detection algorithms that work well in urban driving, do not necessarily work well in an agricultural setting.
Ground plane assumptions common for 3D sensors may break down when applied on rough terrain or high grass.
And vision-based detection algorithms may fail when faced with visually camouflaged objects such as animals and vegetation typical in a natural environment.

In this paper, we present a flexible, multi-modal sensing platform and a dataset for obstacle detection in agriculture.
The platform is mounted on a tractor and includes stereo camera, thermal camera, web camera, 360-degree camera, lidar, and radar. Precise localization is further available from fused IMU and GNSS.
The dataset includes approximately 2 hours of recordings from a grass mowing scenario in Denmark, October 2016.
Both static and moving obstacles are present including humans, mannequin dolls, rocks, barrels, buildings, vehicles, and vegetation.
Ground truth positions of all obstacles were recorded with a drone during operation and have subsequently been manually labeled and synchronized with all sensor data.
The dataset can be downloaded from \url{https://vision.eng.au.dk/fieldsafe/}.

%------------------------------------------------

\begin{figure*}[t]
    \centering
    \includegraphics[height=6cm]{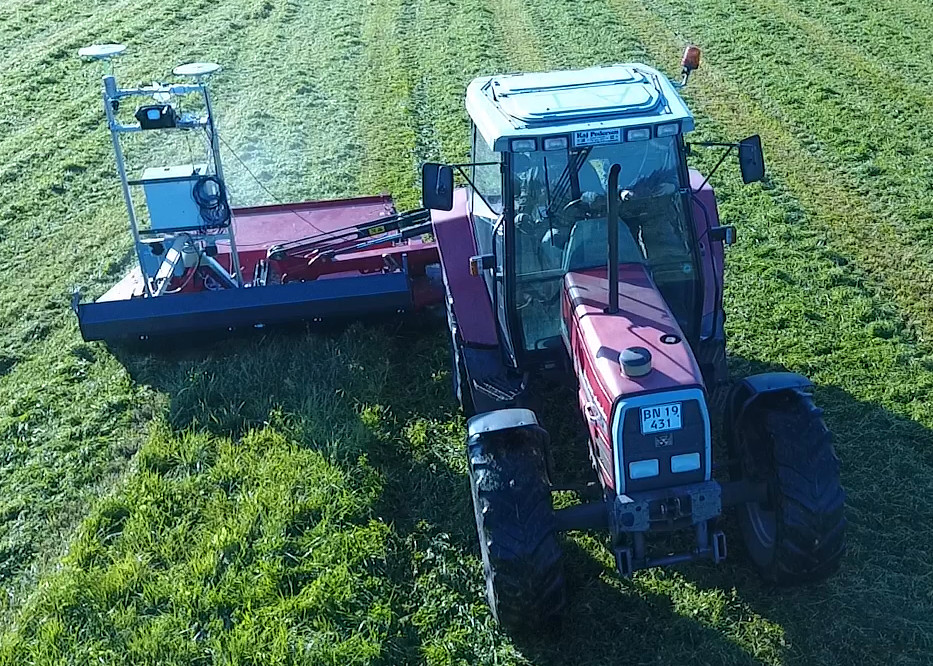}
    \includegraphics[height=6cm]{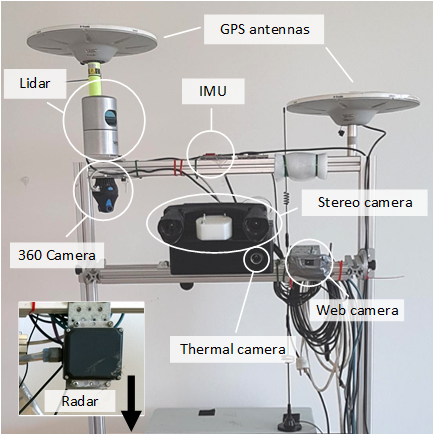}
    \caption{Recording platform.}\label{fig:sensor_platform}
\end{figure*}

\newcommand{\degree}[0]{$^{\circ}$}
\begin{table*}[!htb]
\centering
\caption{Exteroceptive sensors.}
\label{table:sensors_exteroceptive}
\begin{tabular}{|l|l|l|l|l|l|}
\hline
\multicolumn{1}{|c|}{\textbf{Sensor}} & \multicolumn{1}{c|}{\textbf{Model}}                              & \textbf{Resolution} & \multicolumn{1}{c|}{\textbf{FOV}}                                       & \textbf{Range}                                           & \textbf{Data rate} \\ \hline
Stereo camera                         & \begin{tabular}[c]{@{}l@{}}Multisense S21\\ CMV2000\end{tabular} & 1024 x 544          & 85\degree x 50\degree                                                               & 1.5-50m                                                   & 10 fps             \\ \hline
Web camera                            & Logitech HD Pro C920                                             & 1920 x 1080         & 70\degree x 43\degree                                                               & -                                                        & 20 fps             \\ \hline
360-degree camera                            & Giroptic 360cam                                                  & 2048 x 833         & 360\degree x 292\degree                                                             & -                                                        & 30 fps             \\ \hline
Thermal camera                        & Flir A65, 13 mm lens                                             & 640 x 512           & 45\degree x 37\degree                                                               & -                                                        & 30 fps             \\ \hline
Lidar                                 & Velodyne HDL-32E                                                 & 2172 x 32           & 360\degree x 40\degree                                                              & 1-100 m                                                  & 10 fps             \\ \hline
Radar                                 & Delphi ESR                                                       & 32 targets/frame    & \begin{tabular}[c]{@{}l@{}}90\degree x 4.2\degree\\ 20\degree x 4.2\degree\end{tabular} & \begin{tabular}[c]{@{}l@{}}0-60 m\\ 0-174 m\end{tabular} & 20 fps             \\ \hline
\end{tabular}
\end{table*}

\begin{table*}[!htb]
\centering
\caption{Proprioceptive sensors.}
\label{table:sensors_proprioceptive}
\begin{tabular}{|l|l|l|l|l|l|}
\hline
\multicolumn{1}{|c|}{\textbf{Sensor}} & \multicolumn{1}{c|}{\textbf{Model}}  & \textbf{Description} \\ \hline
GPS & Trimble BD982 GNSS & \begin{minipage}[t]{\columnwidth}Dual antenna RTK GPS system. Measures position and horizontal heading of the platform.\end{minipage} \\ \hline
IMU & Vectornav VN-100 & \begin{minipage}[t]{\columnwidth}Measures acceleration, angular velocity, magnetic field, and barometric pressure.\end{minipage}
          \\ \hline
\end{tabular}
\end{table*}

\section{Sensor Setup}
\begin{figure*}[!htb]
    \centering
    \begin{subfigure}[b]{0.45\textwidth}
        \includegraphics[width=\textwidth]{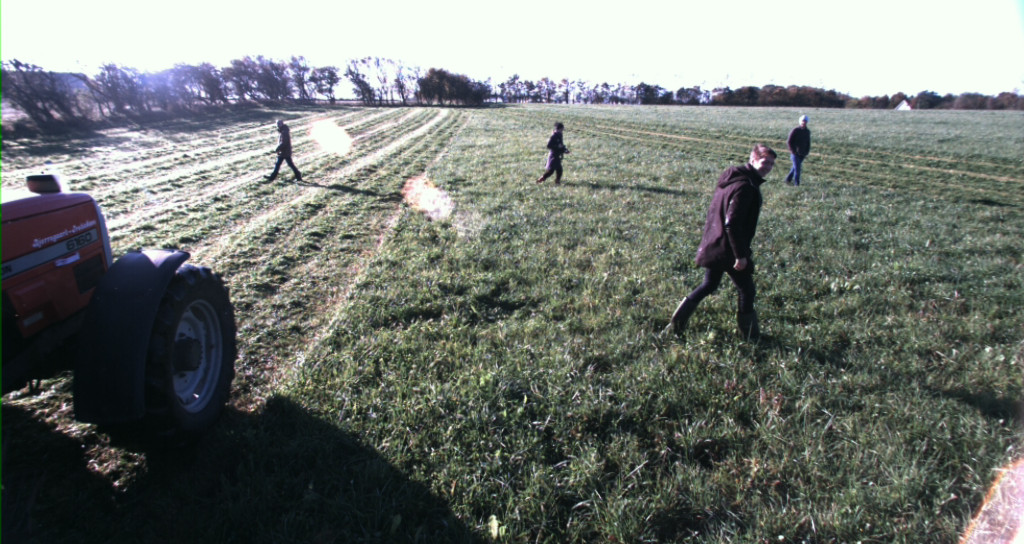}
        \caption{Stereo image}
        \label{fig:stereo_image}
    \end{subfigure}
    ~ %add desired spacing between images, e. g. ~, \quad, \qquad, \hfill etc. 
      %(or a blank line to force the subfigure onto a new line)
    \begin{subfigure}[b]{0.45\textwidth}
        \includegraphics[width=\textwidth]{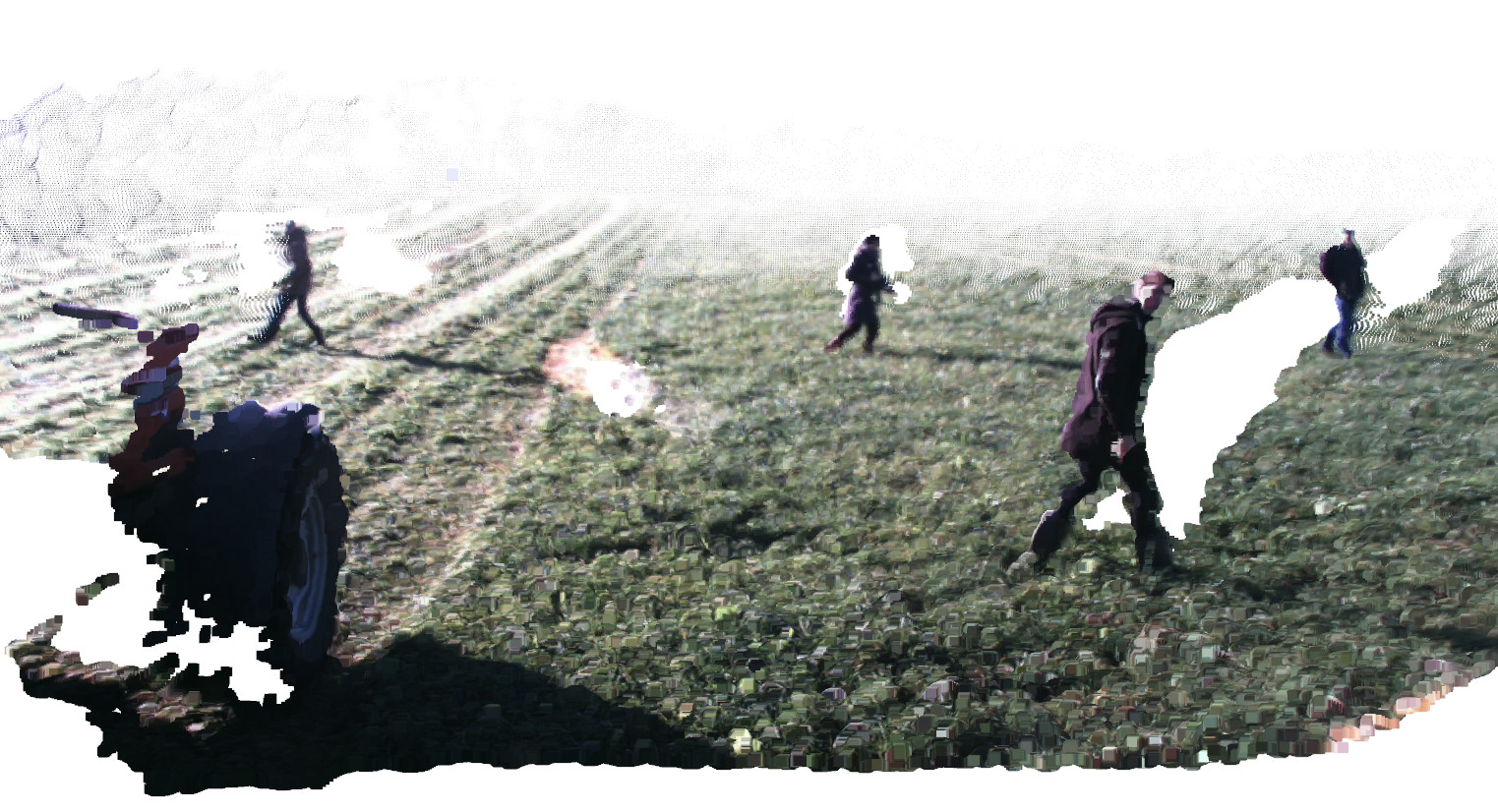}
        \caption{Stereo pointcloud}
        \label{fig:stereo_pointcloud}
    \end{subfigure}
    
    \begin{subfigure}[b]{0.9\textwidth}
        \includegraphics[clip,trim={0 7cm 0 0},width=\textwidth]{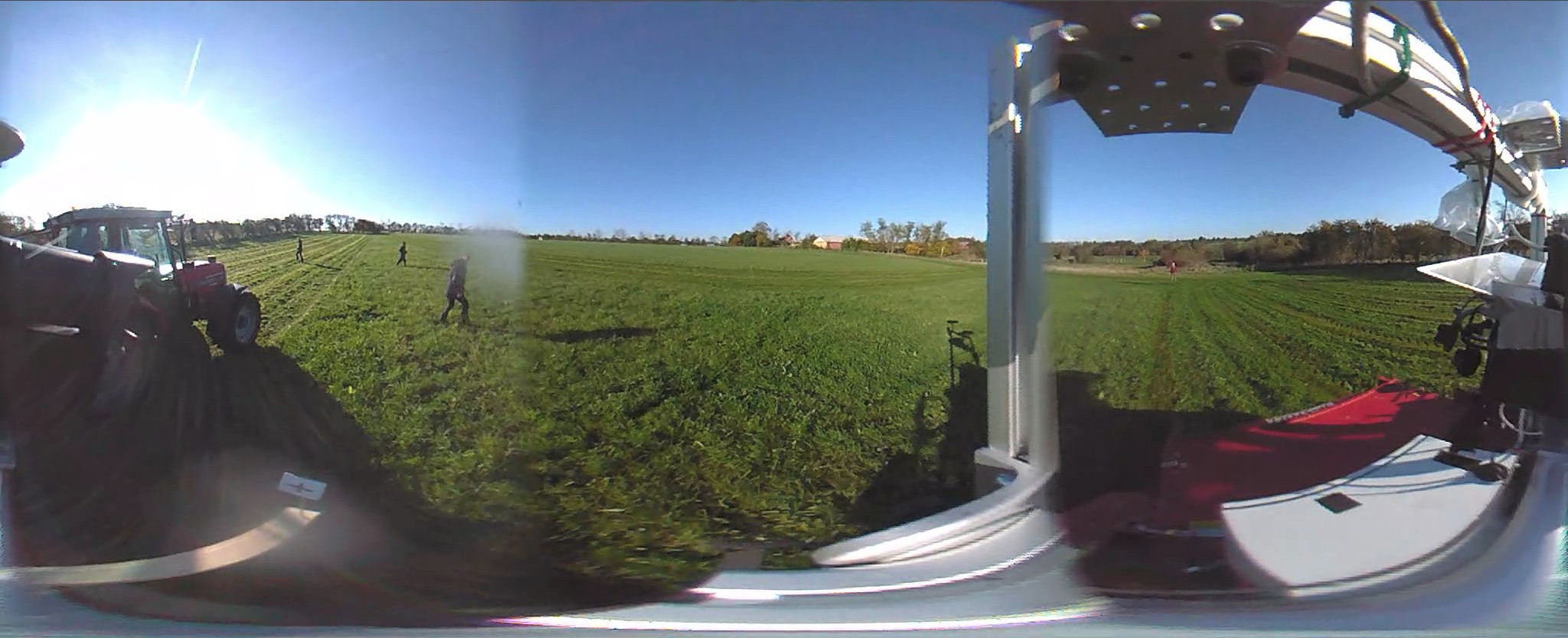}
        \caption{360-degree camera image (cropped)}
        \label{fig:cam360_image}
    \end{subfigure}
    
    \begin{subfigure}[b]{0.45\textwidth}
        \includegraphics[width=\textwidth]{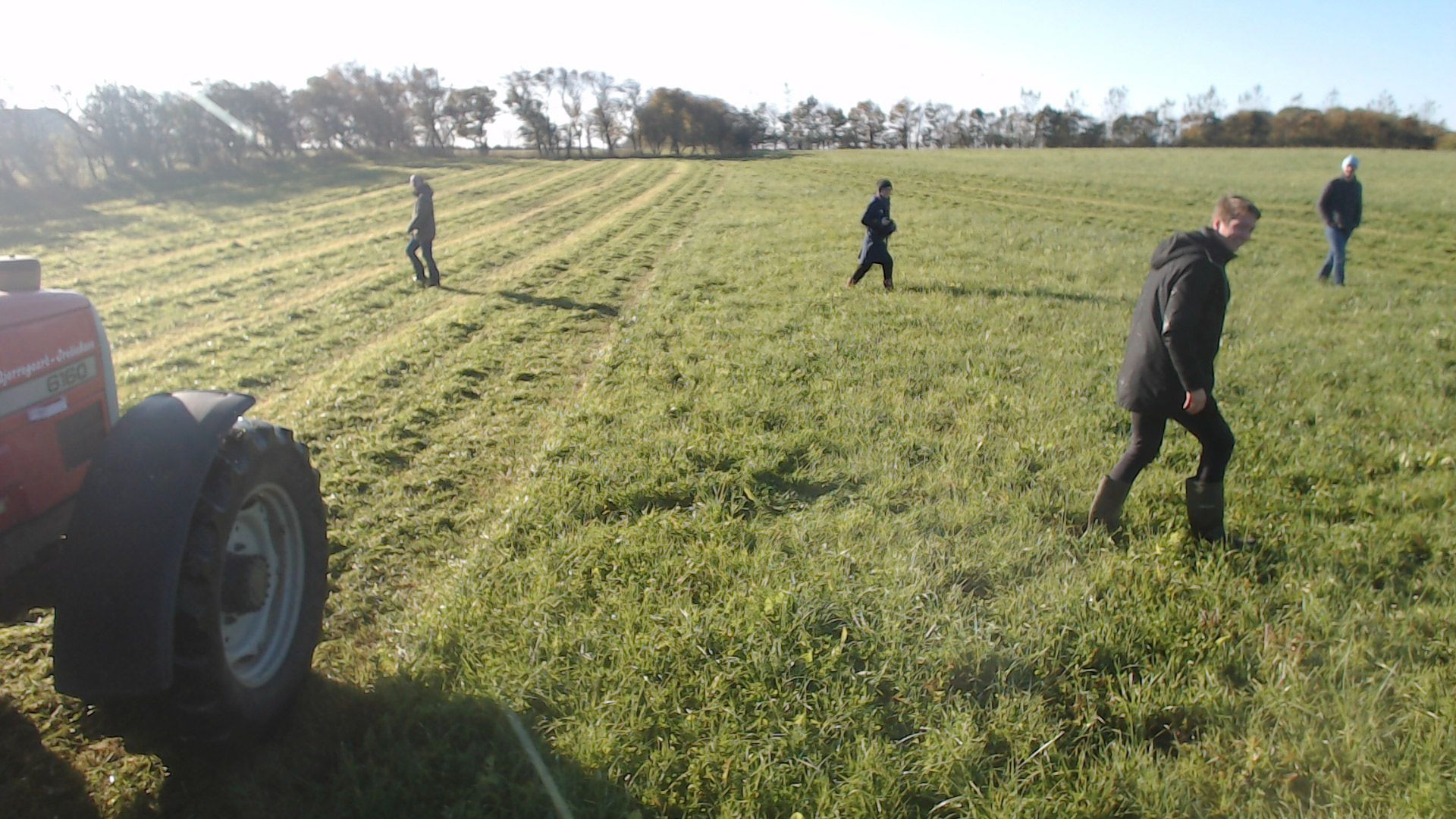}
        \caption{Web camera image}
        \label{fig:webcam_image}
    \end{subfigure}
    \begin{subfigure}[b]{0.45\textwidth}
        \includegraphics[clip,trim={0 5cm 0 0.3cm},width=\textwidth]{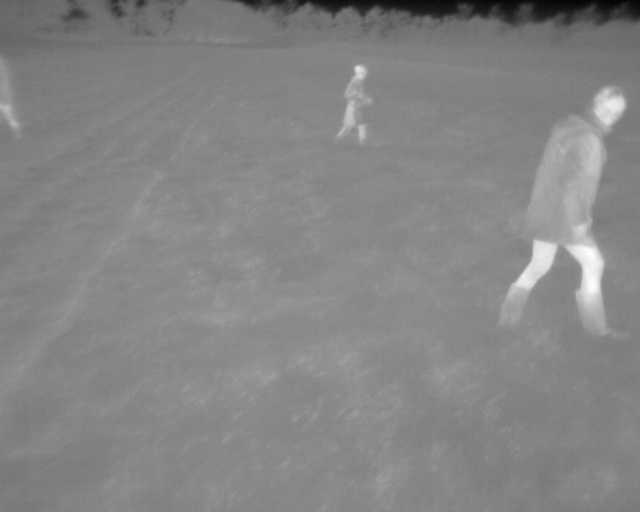}
        \caption{Thermal camera image (cropped)}
        \label{fig:thermal_image}
    \end{subfigure}
    
    \begin{subfigure}[b]{0.9\textwidth}
        \includegraphics[width=\textwidth]{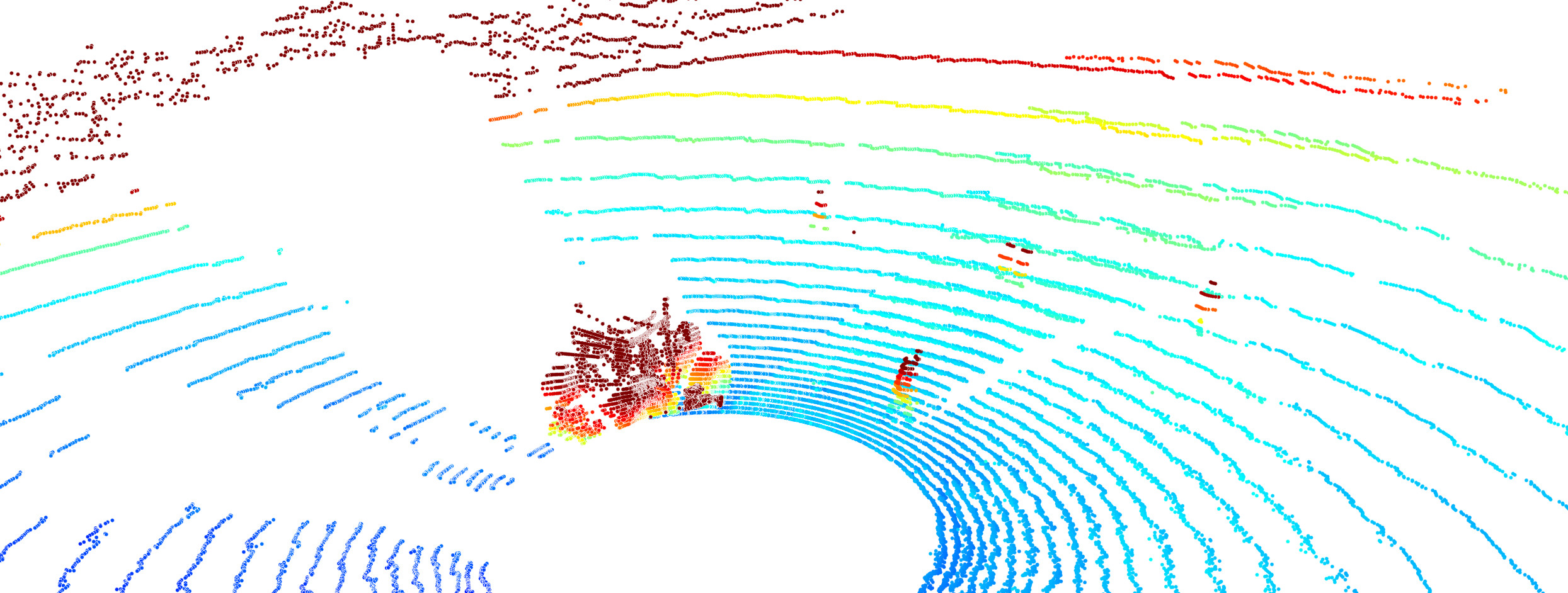}
        \caption{Lidar point cloud (cropped and colored by height)}
        \label{fig:lidar_pointcloud}
    \end{subfigure}
    %\begin{subfigure}[b]{0.3\textwidth}
    %    \includegraphics[width=\textwidth]{Figures/FlirA65_image_raw0054_converted.png}
    %    \caption{Radar targets?}
    %    \label{fig:mouse}
    %\end{subfigure}
    \caption{Example frames from the FieldSAFE dataset.}\label{fig:raw_data}
\end{figure*}

Figure \ref{fig:sensor_platform} shows the recording platform mounted on a tractor during grass mowing.
The platform consists of the exteroceptive sensors listed in Table \ref{table:sensors_exteroceptive}, the proprioceptive sensors listed in Table \ref{table:sensors_proprioceptive}, and a Conpleks Robotech Controller 701 controller used for data collection with the Robot Operating System (ROS) \cite{ROS2009}.
Figure \ref{fig:raw_data} illustrates a synchronized pair of frames from stereo camera, 360-degree camera, web camera, thermal camera, and lidar.
\\
\\
%It is mounted with rubber suspensions to minimize the influence of vibrations on image data.
\textbf{Synchronization}.
Trigger signals for the stereo and thermal cameras were synchronized and generated from a PPS signal from an internal GNSS in the lidar, which allowed exact timestamps for all three sensors.
The remaining sensors were synchronized in software using a best effort approach in ROS.
\\
\\
\textbf{Registration}.
The lidar and the stereo camera were registered with Iterative Closest Point (ICP) as an average over multiple static scenes.
The stereo and thermal cameras were registered using a custom made visual-thermal checkerboard.
The remaining sensors were registered by hand, by estimating extrinsic parameters of their positions.
For a more detailed description, we refer the reader to \cite{Christiansen2017}.

% Example frames from 2016-10-25-11-41-21.bag, timestamp: 1477388888.261 (406.645 s)

%------------------------------------------------

\section{Dataset}
\begin{figure}
    \centering
    \begin{subfigure}[]{0.45\textwidth}
        \includegraphics[width=\textwidth]{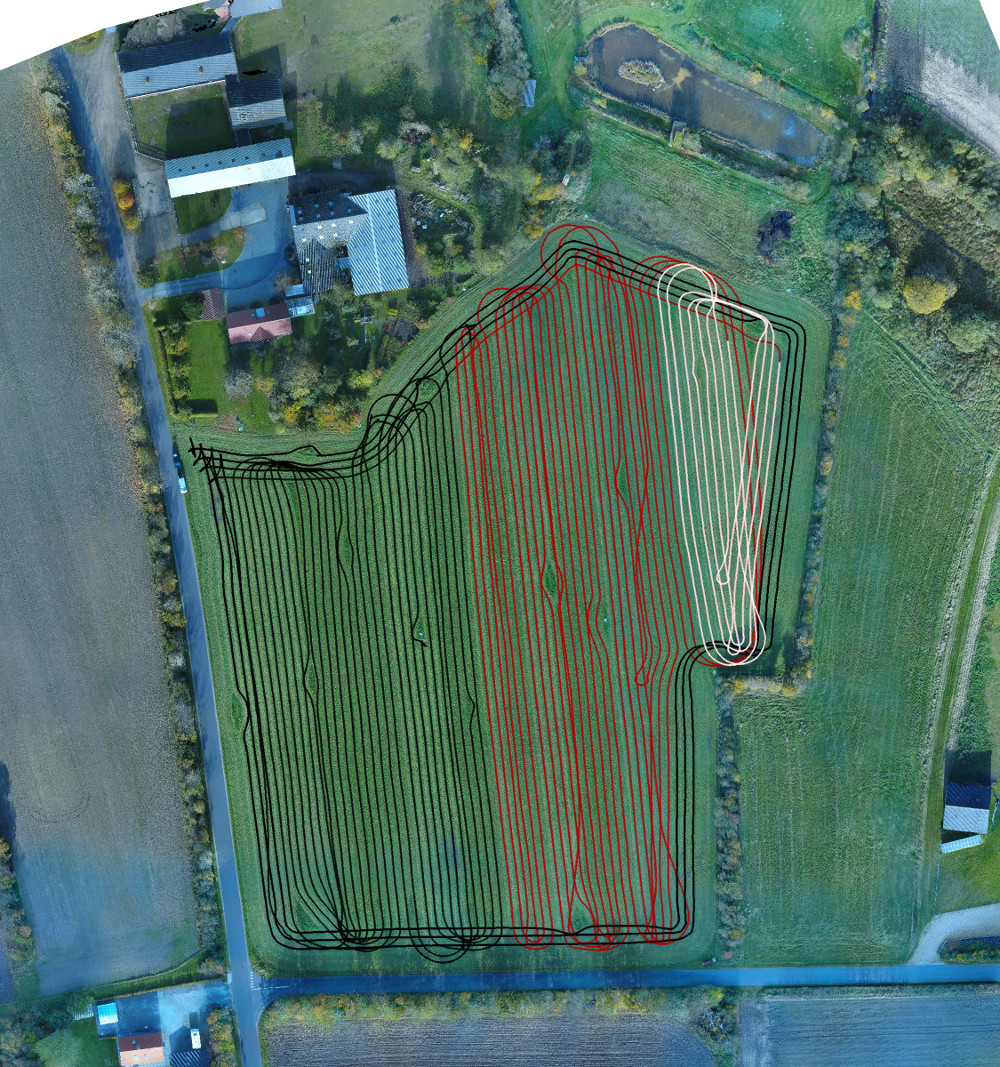}
        \caption{Orthophoto with tractor tracks overlaid. Black tracks include only static obstacles, whereas red and white tracks also have moving obstacles. Currently, red tracks have no ground truth for moving obstacles annotated.}
        \label{fig:orthophoto}
    \end{subfigure}
    \begin{subfigure}[]{0.45\textwidth}
        \includegraphics[width=\textwidth]{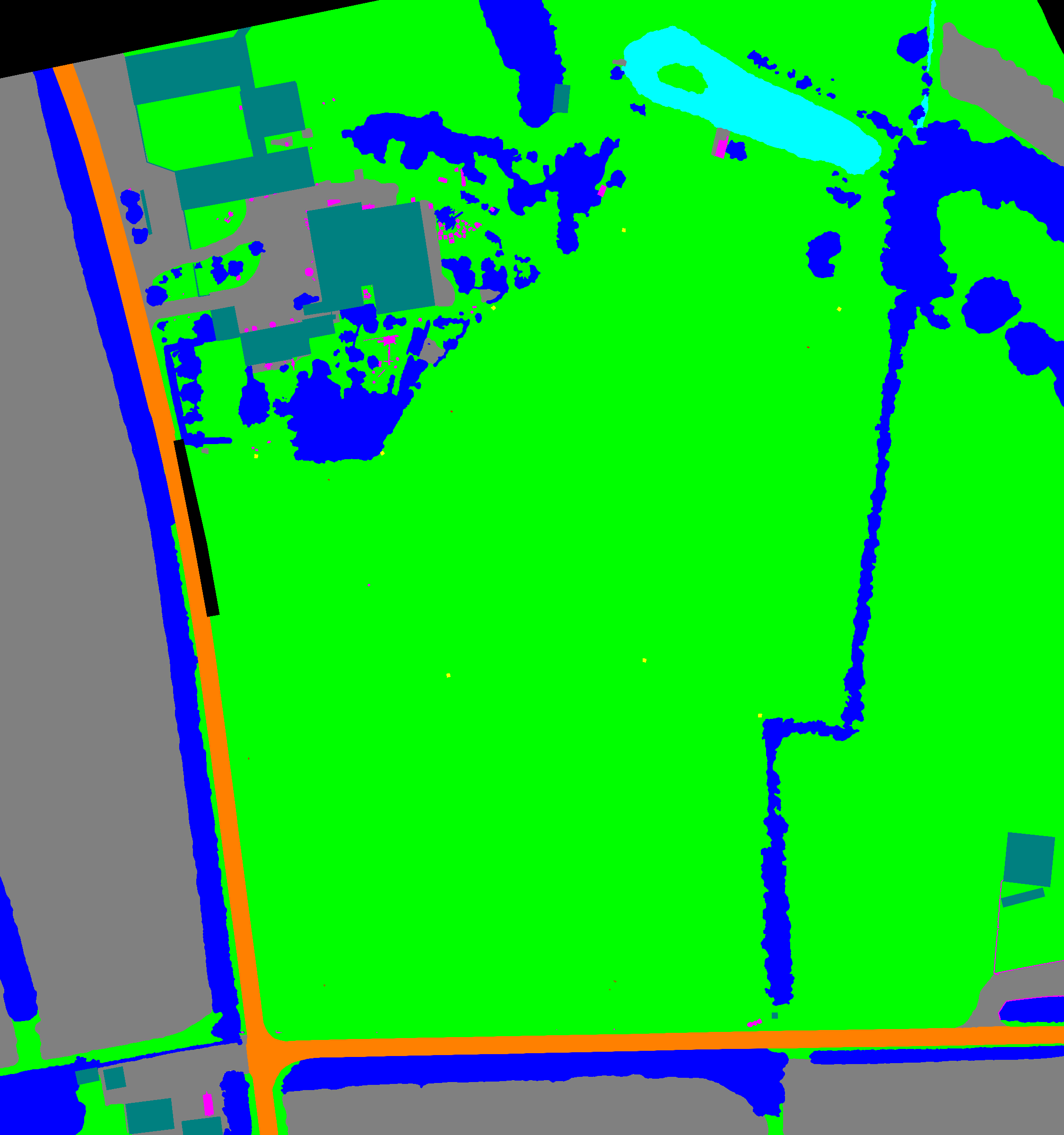}\llap{\includegraphics[width=\textwidth]{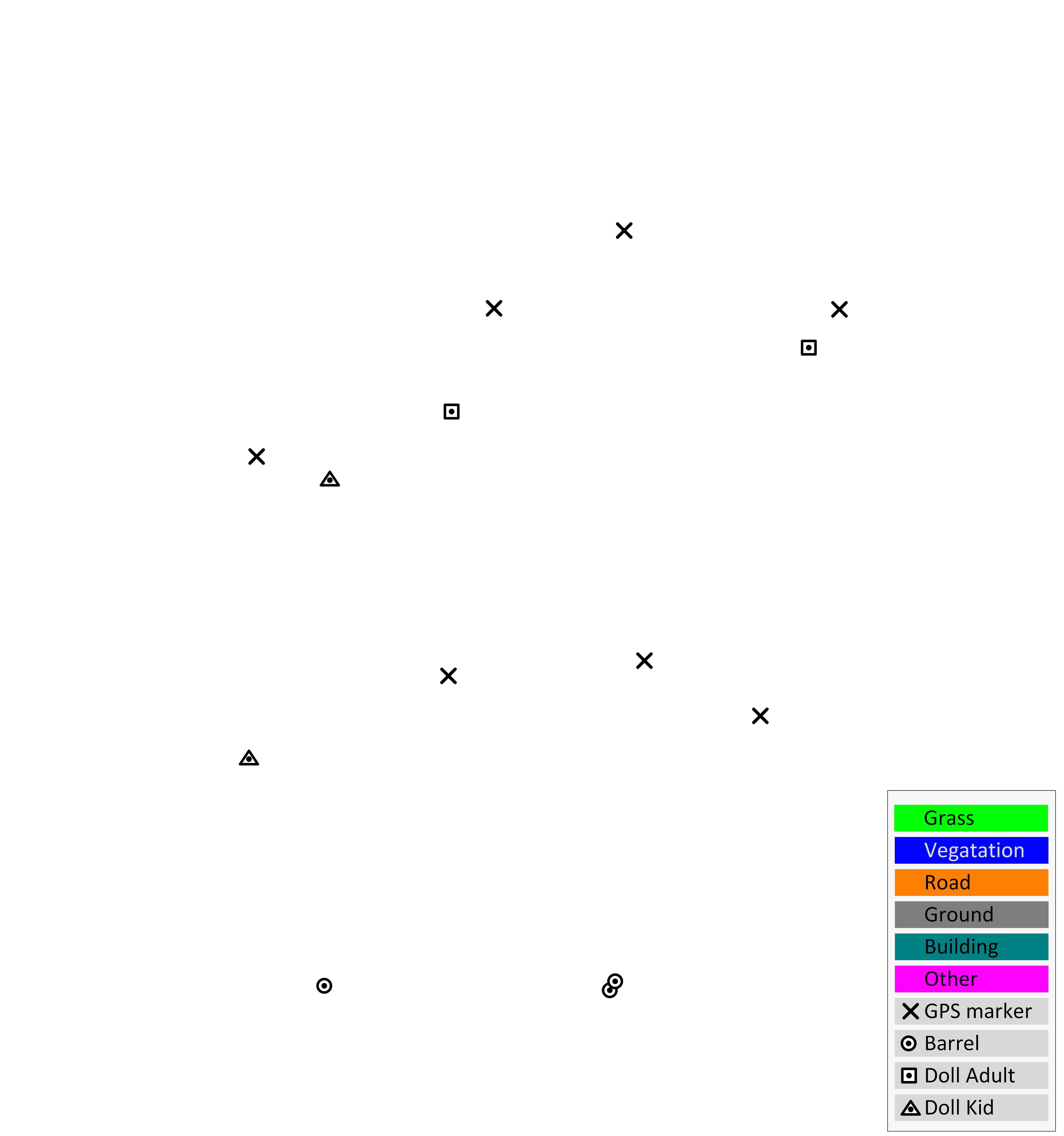}}
        %\leavevmode\makebox(0,0){\put(0,0){\includegraphics[width=\textwidth]{Figures/FieldTrial_D6_Overview.png}}}
        \caption{Labeled orthophoto}
        \label{fig:groundtruth}
    \end{subfigure}
    \caption{Colored and labeled orthophotos.}\label{fig:maps}
\end{figure}

The dataset consists of approximately 2 hours of recordings during grass mowing in Denmark, October 25th 2016.
The exact position of the field was 56.066742, 8.386255 (latitude, longitude).
Figure \ref{fig:orthophoto} shows a map of the field with tractor paths overlaid.
The field is 3.3 ha and surrounded by roads, shelterbelts, and a private property.

\begin{figure*}[t]
    \centering
    \includegraphics[height=3.5cm]{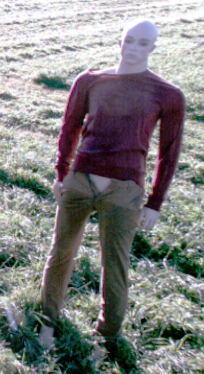}
    \includegraphics[height=3.5cm]{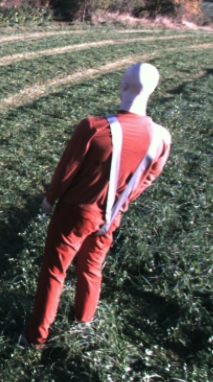}
    \includegraphics[height=3.5cm]{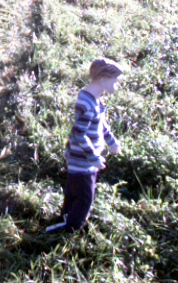}
    \includegraphics[height=3.5cm]{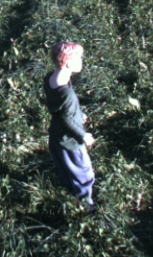}
    \includegraphics[clip,trim={0.8cm 0 1.2cm 0},height=3.5cm]{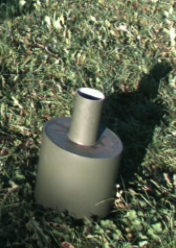}
    \includegraphics[clip,trim={1cm 0 1cm 0},height=3.5cm]{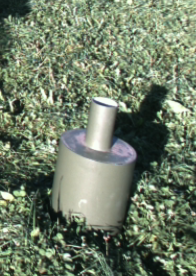}
    \includegraphics[height=3.5cm]{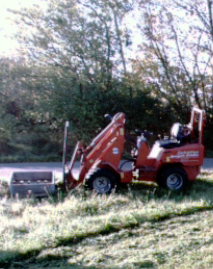}
    \caption{Examples of static obstacles.}\label{fig:static_obstacles}
\end{figure*}

\begin{figure*}[!htb]
    \centering
    \begin{subfigure}[b]{0.20\textwidth}
        \includegraphics[width=3.0cm]{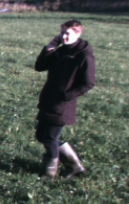}\\
        \includegraphics[width=3.0cm]{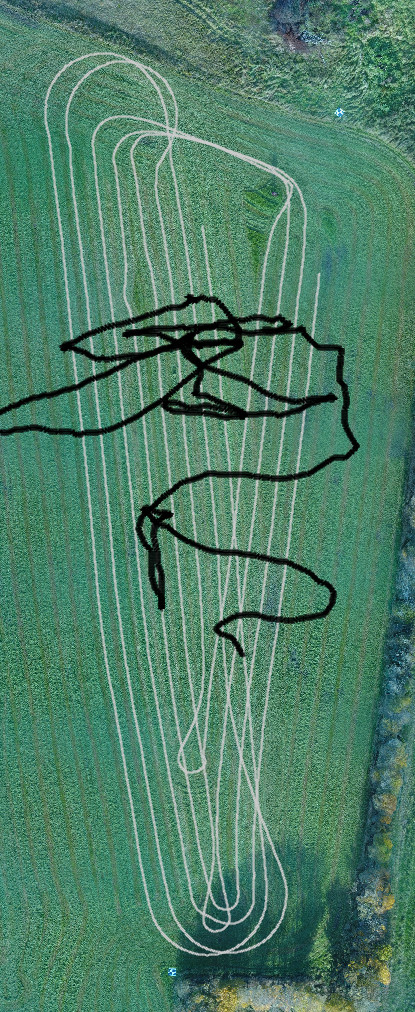}
        \caption{Human 1}
        \label{fig:human1}
    \end{subfigure}
    \begin{subfigure}[b]{0.20\textwidth}
        \includegraphics[width=3.0cm]{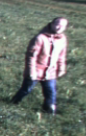}\\
        \includegraphics[width=3.0cm]{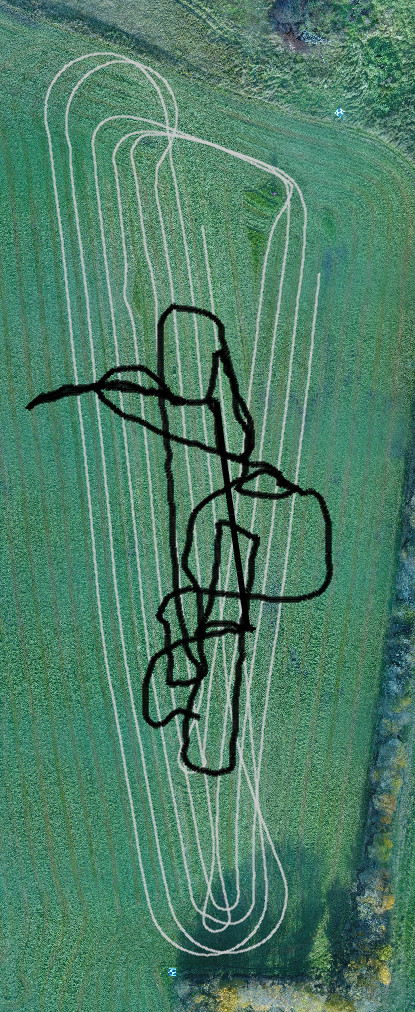}
        \caption{Human 2}
        \label{fig:human2}
    \end{subfigure}
    \begin{subfigure}[b]{0.20\textwidth}
        \includegraphics[width=3.0cm]{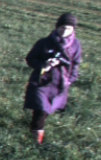}\\
        \includegraphics[width=3.0cm]{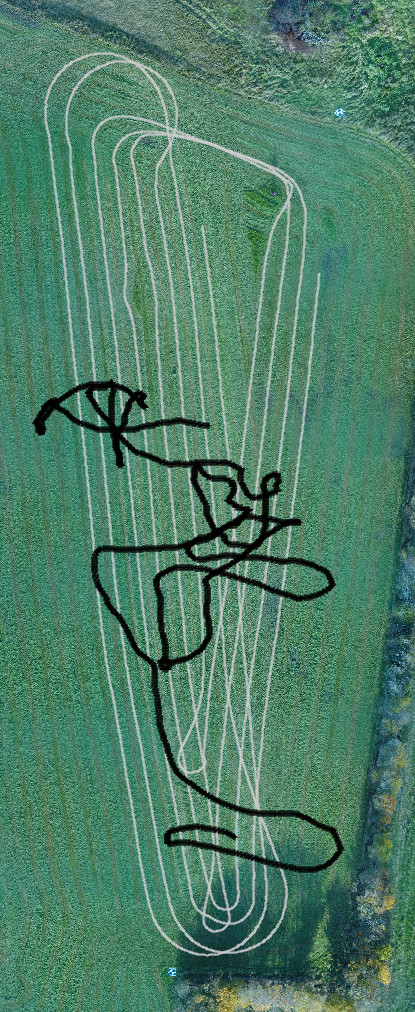}
        \caption{Human 3}
        \label{fig:human3}
    \end{subfigure}
    \begin{subfigure}[b]{0.20\textwidth}
        \includegraphics[width=3.0cm]{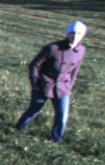}\\
        \includegraphics[width=3.0cm]{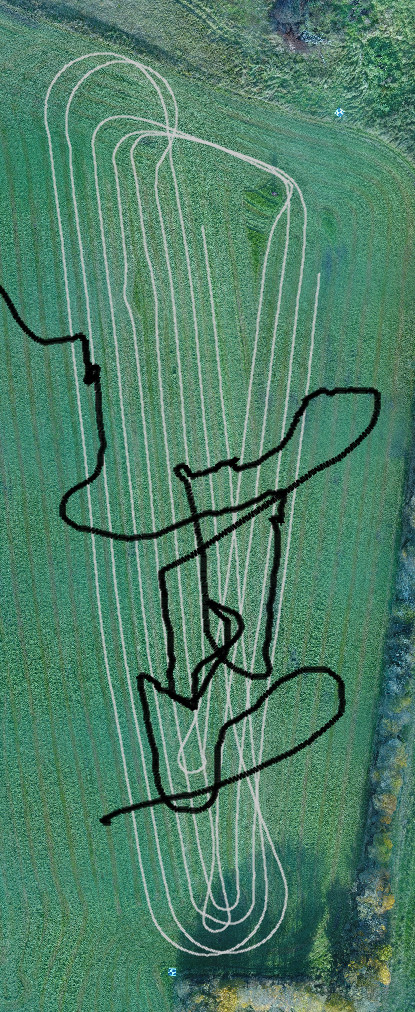}
        \caption{Human 4}
        \label{fig:human4}
    \end{subfigure}
    \caption{Examples of moving obstacles (from the stereo camera) and their paths (black) overlaid on tractor path (grey).}\label{fig:dynamic_obstacles}
\end{figure*}

A number of static obstacles exemplified in Figure \ref{fig:static_obstacles} were placed on the field prior to recording.
They included mannequin dolls (adults and children), rocks, barrels, buildings, vehicles, and vegetation.
Figure \ref{fig:groundtruth} shows the placement of static obstacles on the field overlaid on a ground truth map colored by object classes.
%Furthermore, a number of obstacles were present in a small residential area next to the field and thus visible to most sensors.

Additionally, a session with moving obstacles was recorded where four humans were told to walk in random patterns.
Figure \ref{fig:dynamic_obstacles} shows the four subjects and their respective paths on a subset of the field.
The subset corresponds to the white tractor tracks in Figure \ref{fig:orthophoto}.
The humans crossed the path of the tractor a number of times, thus emulating dangerous situations that must be detected by a safety system.
Along the way, various poses such as standing, sitting, and lying were represented.

During the entire traversal and mowing of the field, data from all sensors were recorded.
Along with video from a hovering drone, a static orthophoto from another drone, and corresponding manually annotated class labels, these are all available from the FieldSAFE website.

%------------------------------------------------

\section{Ground Truth}
Ground truth information on object location and class labels for both static and moving obstacles is available as timestamped global (geographic) coordinates.
By transforming local sensor data from the tractor into global coordinates, a simple look-up of class label into the annotated ground truth map is possible.

Prior to traversing and mowing the field, a number of custom-made markers were distributed on the ground and measured with exact global coordinates using a handheld Topcon GRS-1 RTK GNSS.
A DJI Phantom 4 drone was used to take overlapping bird's-eye view images of an area covering the field and its surroundings.
Pix4D \cite{Pix4D} was used to stitch the images and generate a high-resolution orthophoto (Figure \ref{fig:orthophoto}) with a ground sampling distance (GSD) of 2 cm.
The orthophoto was manually labeled pixel-wise as either \textit{grass}, \textit{ground}, \textit{road}, \textit{vegetation}, \textit{building}, \textit{GPS marker}, \textit{barrel}, \textit{human}, or \textit{other} (Figure \ref{fig:groundtruth}).
Using the GPS coordinates of the markers and their corresponding positions in the orthophoto, a mapping between GPS coordinates and pixel coordinates was estimated.

For annotating the location of moving obstacles, a DJI Matrice 100 was used to hover approximately 75 m above the ground while the tractor traversed the field.
The drone recorded video at 25 fps with a resolution of 1920x1080.
Due to limited battery capacity, the recording was split into two sessions of each 20 minutes.
The videos were manually synchronized with sensor data from the tractor by introducing physical synchronization events in front of the tractor in the beginning and end of each session. 
Using the 7 GPS markers that were visible within field of view of the drone, the videos were stabilized and warped to a bird's-eye view of a subset of the field.
As described above for the static orthophoto, GPS coordinates of the markers and their corresponding positions in the videos were then used to generate a mapping between GPS coordinates and pixel coordinates.
Finally, the moving obstacles were manually annotated in each frame of one of the videos using the vatic video annotation tool \cite{vondrick2013efficiently}.
Figure \ref{fig:dynamic_obstacles} shows the path of each object overlaid on a subset of the orthophoto.
The second video is yet to be annotated.

%------------------------------------------------

\section{Summary and Future Work}
In this paper, we have presented a calibrated and synchronized multi-modal dataset for obstacle detection in agriculture.
We envision the dataset to facilitate a wide range of future research within autonomous agriculture and obstacle detection for farming vehicles.

In future work, we plan on annotating the remaining session with moving obstacles.
Additionally, we would like to extend the dataset with more scenarios from various agricultural environments while widening the range of encountered illumination and weather conditions.

Currently, all annotations reside in a global coordinate system.
Projecting these annotations to local sensor frames inevitably causes localization errors.
Therefore, we would like to extend annotations with e.g object bounding boxes for each sensor.
\\
\\

%----------------------------------------------------------------------------------------
%	REFERENCE LIST
%----------------------------------------------------------------------------------------

\bibliographystyle{plain}
\bibliography{references}

\begin{thebibliography}{10}

\bibitem{Didi1}
{Didi Data Release \#2 - Round 1 Test Sequence and Training}.

\bibitem{Abidine2004}
Aziz~Z. Abidine, Brian~C. Heidman, Shrini~K. Upadhyaya, and David~J. Hills.
\newblock {Autoguidance system operated at high speed causes almost no tomato
  damage}.
\newblock {\em California Agriculture}, 58(1):44--47, jan 2004.

\bibitem{AutonomousSolutions}
{ASI}.
\newblock {Autonomous Solutions}.
\newblock \url{https://www.asirobots.com/farming/}, 2016.
\newblock Accessed: 2017-08-09.

\bibitem{TMEMotorwayDataset}
Claudio Caraffi, Tomas Vojir, Jura Trefny, Jan Sochman, and Jiri Matas.
\newblock {A System for Real-time Detection and Tracking of Vehicles from a
  Single Car-mounted Camera}.
\newblock In {\em ITS Conference}, pages 975--982, Sep. 2012.

\bibitem{CaseIH}
{Case IH}.
\newblock {Case IH Autonomous Concept Vehicle}.
\newblock
  \url{http://www.caseih.com/apac/en-in/news/pages/2016-case-ih-premieres-concept-vehicle-at-farm-progress-show.aspx},
  2016.
\newblock Accessed: 2017-08-09.

\bibitem{Christiansen2017}
P.~Christiansen, M.~Kragh, K.~A. Steen, H.~Karstoft, and R.~N. J{\o}rgensen.
\newblock Platform for evaluating sensors and human detection in autonomous
  mowing operations.
\newblock {\em Precision Agriculture}, 18(3):350--365, Jun 2017.

\bibitem{Cordts2016Cityscapes}
Marius Cordts, Mohamed Omran, Sebastian Ramos, Timo Rehfeld, Markus Enzweiler,
  Rodrigo Benenson, Uwe Franke, Stefan Roth, and Bernt Schiele.
\newblock The cityscapes dataset for semantic urban scene understanding.
\newblock In {\em Proc. of the IEEE Conference on Computer Vision and Pattern
  Recognition (CVPR)}, 2016.

\bibitem{DIPLECS}
{DIPLECS}.
\newblock {DIPLECS Autonomous Driving Datasets}.
\newblock \url{http://ercoftac.mech.surrey.ac.uk/data/diplecs/}, 2015.
\newblock Accessed: 2017-08-31.

\bibitem{GaidonVirtualCVPR2016}
A~Gaidon, Q~Wang, Y~Cabon, and E~Vig.
\newblock Virtual worlds as proxy for multi-object tracking analysis.
\newblock In {\em CVPR}, 2016.

\bibitem{Geiger2013IJRR}
Andreas Geiger, Philip Lenz, Christoph Stiller, and Raquel Urtasun.
\newblock Vision meets robotics: The kitti dataset.
\newblock {\em International Journal of Robotics Research (IJRR)}, 2013.

\bibitem{insight}
{InSight}.
\newblock {InSight SHRP2}.
\newblock \url{https://insight.shrp2nds.us/}, 2017.
\newblock Accessed: 2017-08-31.

\bibitem{amuse2013gt}
Philipp Koschorrek, Tommaso Piccini, Per \"{O}berg, Michael Felsberg, Lars
  Nielsen, and Rudolf Mester.
\newblock A multi-sensor traffic scene dataset with omnidirectional video.
\newblock In {\em Ground Truth - What is a good dataset? CVPR Workshop 2013},
  2013.

\bibitem{Kubota}
{Kubota}.
\newblock {Kubota}.
\newblock \url{http://www.kubota-global.net/news/2017/20170125.html}, 2017.
\newblock Accessed: 2017-08-16.

\bibitem{RobotCarDatasetIJRR}
Will Maddern, Geoff Pascoe, Chris Linegar, and Paul Newman.
\newblock {1 Year, 1000km: The Oxford RobotCar Dataset}.
\newblock {\em The International Journal of Robotics Research (IJRR)},
  36(1):3--15, 2017.

\bibitem{MatzenICCV13}
Kevin Matzen and Noah Snavely.
\newblock Nyc3dcars: A dataset of 3d vehicles in geographic context.
\newblock In {\em Proc. Int. Conf. on Computer Vision}, 2013.

\bibitem{MVD2017}
Gerhard Neuhold, Tobias Ollmann, Samuel Rota~Bul\`o, and Peter Kontschieder.
\newblock The mapillary vistas dataset for semantic understanding of street
  scenes.
\newblock In {\em International Conference on Computer Vision (ICCV)}, 2017.

\bibitem{Pix4D}
{Pix4D}.
\newblock {Pix4D}.
\newblock \url{http://pix4d.com/}, 2014.
\newblock Accessed: 2017-09-05.

\bibitem{ROS2009}
Morgan Quigley, Ken Conley, Brian~P. Gerkey, Josh Faust, Tully Foote, Jeremy
  Leibs, Rob Wheeler, and Andrew~Y. Ng.
\newblock Ros: an open-source robot operating system.
\newblock In {\em ICRA Workshop on Open Source Software}, 2009.

\bibitem{RosCVPR16}
German Ros, Laura Sellart, Joanna Materzynska, David Vazquez, and Antonio
  Lopez.
\newblock {The SYNTHIA Dataset}: A large collection of synthetic images for
  semantic segmentation of urban scenes.
\newblock 2016.

\bibitem{DidiChallenge}
Udacity and Didi.
\newblock {Udacity Didi \$100k Challenge Dataset 1}.

\bibitem{Didi2}
Inc. Udacity.
\newblock {Udacity Didi Challenge - Round 2 Dataset}.

\bibitem{vondrick2013efficiently}
Carl Vondrick, Donald Patterson, and Deva Ramanan.
\newblock Efficiently scaling up crowdsourced video annotation.
\newblock {\em International Journal of Computer Vision}, 101(1):184--204,
  2013.

\end{thebibliography}

%\begin{thebibliography}{99} % Bibliography - this is intentionally simple in this template
%
%\bibitem[Figueredo and Wolf, 2009]{Figueredo:2009dg}
%Figueredo, A.~J. and Wolf, P. S.~A. (2009).
%\newblock Assortative pairing and life history strategy - a cross-cultural
%  study.
%\newblock {\em Human Nature}, 20:317--330.
% 
%\end{thebibliography}

%----------------------------------------------------------------------------------------

\end{document}